\documentclass[letterpaper, 10 pt, conference]{ieeeconf}  

\IEEEoverridecommandlockouts                              

\usepackage{amsfonts, amsmath}
\usepackage[utf8]{inputenc}
\usepackage{caption}
\usepackage{subcaption}
\usepackage{algorithm}
\usepackage{bbm}
\usepackage{graphicx}
\usepackage{tikz}
\usepackage{wrapfig}
\usepackage{float}
\usepackage{algpseudocode}
\usepackage{geometry} 

\graphicspath{ {./images/} }

\newtheorem{defn}{Definition}[section]
\newtheorem{prop}{Proposition}[section]

\usepackage[normalem]{ulem} 

\DeclareMathOperator*{\argmax}{arg\,max}

\title{\LARGE \bf
Efficient and robust Sensor Placement in Complex Environments
}

\author{Lukas Taus$^{1}$ and Yen-Hsi Richard Tsai$^{2}$
\thanks{$^{1,2}$Oden Institute for Computational Engineering and Sciences, The University of Texas at Austin}
\thanks{{\tt\small $^{1}$l.taus@utexas.edu}}
\thanks{{\tt\small $^{2}$ytsai@math.utexas.edu}}%
}

\begin{document}

\maketitle
\thispagestyle{empty}
\pagestyle{empty}

\begin{abstract}
We address the problem of efficient and unobstructed surveillance or communication in complex environments. On one hand, one wishes to use a minimal number of sensors to cover the environment. On the other hand, it is often important to consider solutions that are robust against sensor failure or adversarial attacks. This paper addresses these challenges of designing minimal sensor sets that achieve multi-coverage constraints --- every point in the environment is covered by a prescribed number of sensors.  
We propose a greedy algorithm to achieve the objective. Further, we explore deep learning techniques to accelerate the evaluation of the objective function formulated in the greedy algorithm. The training of the neural network reveals
that the geometric properties of the data significantly impact the network's performance, particularly at the end stage. By taking into account these properties, we discuss the differences in using greedy and $\epsilon$-greedy algorithms to generate data and their impact on the robustness of the network.
\end{abstract}

\section{INTRODUCTION}
In this paper, we focus on the problem of  sensor placement in environments with obstacles blocking the sensors' lines of sight. 
The sensors that we have in mind include conventional cameras (passive observations)  and LIDARs or other active sensors that emit and receive high-frequency signals. Depending on the sensor type, the problem is that of surveillance or communication. The goal is to place a minimal number of sensors so that multiple sensors can observe each point in the environment. 
The motivation for such a problem is to improve the robustness of the sensor network against sensor failure and adversarial attacks.
Naturally, the need for solving such type of constrained optimization problems can be found in many practical problems. 
The placement of 5G towers, for example, 
could benefit from optimal solutions to the multi-coverage constraints to improve the robustness of communication against signal loss due to blockage. 

Due to the constraint of requiring multiple sensors to observe the free space in the environment, finding optimal configurations of a minimal set of sensors requires a daunting computational complexity. We propose a greedy approach, accelerated by deep learning, to efficiently generate a sequence of sensor locations that achieve multi-coverage for most free space. We also present numerical experiments illustrating results for real-life urban environments.

\subsection{RELATED WORKS}
Finding optimal sensor locations for a known environment is related to the gallery problem in computational geometry. For polygonal environments with $n$ vertices and $h$ holes it was shown in \cite{BjorlingSachs1995AnEA} \cite{185346} that $\lfloor (n + h) / 3 \rfloor$ sensors are sufficient. However, determining the optimal sensor locations for an environment was shown to be NP-complete in \cite{URRUTIA2000973} \cite{1056648} \cite{1057165}. For more general environments, an alternating minimization scheme \cite{GoroMinProb} and a system of differential equations \cite{kim2016optimal} were proposed. Both methods assume that the number of sensors is fixed. For general 2D environments, Landa et al. \cite{landa2006visibility} \cite{landa2008visibility} \cite{landa2007robotic} propose to place sensors along frontiers, the boundary between the observed and occluded space. However, it is not necessarily optimal to choose sensor locations along the frontiers \cite{StarMapExample}. For general 3D environments, a level set representation is used to calculate visible regions \cite{cheng2005visibility} \cite{tsai2004visibility}. For environments that are given by a graph of a function, a method to construct volumetric visibility information was discussed in \cite{visalg}. A common measure of information gain is the volume of the unexplored region within sensor range \cite{greedy1} \cite{greedy2} \cite{greedy3} \cite{greedy4}. However, other measures like the surface area of frontiers weighted by the view angle \cite{valente2012information} \cite{valente2014information} have also been used. This information can be used in greedy algorithms where a sequence of sensor locations is generated through maximization of the used measure \cite{ly2018greedy}. Using the gained volume of observed space as a gain function and approximating it using deep learning techniques was considered in \cite{explorationsurveilance}. In addition to achieving visibility, it is also indispensable in practice to account for depth-sensor errors by considering uncertainty quantification. Popovi\'c et al. proposed a method to accurately estimate the occlusion map from faulty or incomplete depth measurements \cite{uncertaintypaper}. 

\section{MULTI-COVERAGE FORMULATION}
We consider the task of generating a sequence of sensor locations to achieve visibility in a given environment, using as few sensors as possible. At the same time, we propose that multiple sensors can observe the same region of the environment. This multi-coverage constraint would increase the robustness of surveillance against sensor failure or possible imperfection in the given information about the environment. Even when one sensor fails, it is still possible to observe the full environment. When the sensors are depth sensors, it also provides improved confidence in the depth measurement since we can access multiple measurements to the same surface from different angles. This is especially important when the depth sensors are unreliable due to the surface properties of the environment, e.g. reflective surfaces.

\subsection{The order of visibility}
Consider an environment $\Omega\subset\mathbb{R}^d$, d=2,3. The environment consists of two disjoint sets:  an open set $\Omega_{\text{free}}$ which describes the free space and a closed set $\Omega_{\text{obs}}:=\Omega\setminus\Omega_\text{free}$ which describes the space that is covered by some obstacle.

\begin{figure}[H]
    \centering
    \includegraphics[width = 6cm]{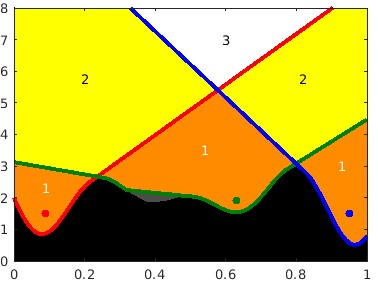}
    \caption{The order of visibility of three sensors (red, blue, and green). Here $\Omega_\text{free}$ is the area above the black curve on the bottom.}\label{fig:vis-order}
\end{figure}

For line-of-sight-coverage, define the relation between any pair of points in $\Omega$:
\begin{equation}\label{def:vis_relation}
  x \overset{\text{vis}}{\sim} y \iff \forall \lambda \in [0,1] \text{ : } \lambda x + (1-\lambda) y \in \Omega_{free}.  
\end{equation}
We define the "order of visibility" map by
\begin{defn}{(Order of visibility map)}
Given an environment $\Omega$ and sensors $x_1,x_2,...,x_n \in \Omega_{\text{free}}$. 
The order of visibility at a point $y \in \Omega_{\text{free}}$ is defined by
$$\mathcal{O}_\text{vis}(y;x_1,x_2,...,x_n) = \sum_{i=1}^n \Phi(y;x_i),$$
where $\Phi(y, x_i)$ is the indicator function of $y \overset{\text{vis}}{\sim} x_i$. 
\end{defn}
Note that two sensors $x_i$ and $x_j$ might be in the same location i.e. $x_i = x_j$. In the following we will also use the notation $\mathcal{O}_\text{vis}(y;P_n)$ for a set of sensors $P_n = \{x_1,...,x_n\}$ for the above. We illustrate the concept of visibility order in Figure~\ref{fig:vis-order}. 

Thus we can describe our objective  as follows:\medskip\\
\fbox{\begin{minipage}{0.97\linewidth}
Find the minimal number of sensors\\$P_n = \{x_1,...,x_n\}$ such that
\begin{equation}
  \small{\text{Vol} \left( \left\{y\in \Omega \vert~ \mathcal{O}_\text{vis}(y; P_n) \geq k \right\} \right) = \delta \text{Vol}(\Omega_{\text{free}})}.  
\end{equation}
\end{minipage}}\hfill \\

\noindent Solving this problem yields a sequence of sensor locations $x_1,...,x_n$ such that a threshold $\delta \in (0,1)$ of the free space of the environment $\Omega_{\text{free}}$ is observed by at least $k$ sensors. \\

Unfortunately, this problem is computationally very complex. Even when the environment is a polygon with $n$ vertices and $h$ holes, calculating optimal locations is an NP-complete problem \cite{URRUTIA2000973} \cite{1056648} \cite{1057165}. Other methods work for more general environments, but the amount of sensors needed is assumed to be known \cite{GoroMinProb} \cite{kim2016optimal}. 

For simplicity, we consider obstacle maps that are given by a graph of a function $f$ via
$$    (x_1,...,x_n) \in \Omega_{\text{free}} \iff f(x_1,...,x_{n-1}) < x_n. $$
Consequently, one can prove that 
If the environment is described by a function $f$. Then 
for any $y \in \Omega_{\text{free}}$ there exists a function $g_y$ such that\\
$y \overset{\text{vis}}{\sim} x = (x_1,...,x_n) \iff g_y(x_1,...,x_{n-1}) < x_n.$
This means that the visibility from the locations $x_1,...,x_n$ can be described by the graph of the function $g_y$.

While this assumption reduces the dimension of the underlying problem, it still allows complicated environment geometries.
Nevertheless, we remark that the proposed greedy approach and the accompanying Deep Learning strategy to be described below do not really depend on this assumption on the environment. 

\section{GREEDY ALGORITHMS}
To solve this problem in a feasible time greedy algorithms are used to calculate approximate solutions. They involve generating a sequence of sensors one by one where the next point is chosen by maximizing some objective function. We call the value of such an objective function at a point $x$, the gain at $x$. For our problem, a natural choice of the objective function would be the gain of observed area up to order $k$. Define for any set of sensors the function
$$\Psi_k(P) = \int_{\Omega_{\text{free}}} \min(\mathcal{O}_\text{vis}(y;P), k) dy.$$
Then the gain function for some given set of sensors $P_m = \{x_1,...,x_m\}$ is defined as 
\begin{equation}
    \mathcal{G}_\text{naïve}(x; P_m):= \Psi_k(P_m \cup \{x\}) - \Psi_k(P_m).
\end{equation}
However, as the following Proposition shows, this naïve choice of objective function will lead to trivial and non-optimal solutions.

\begin{prop}
Suppose $P_m = \{x_1,...,x_m\}$ are generated by 
$$x_{m+1} \in \argmax_{x \in \Omega_{\text{free}}} \mathcal{G}_\text{naïve}(x; P_m)$$
and
$$\forall y \in \{z \in \Omega_{\text{free}} \vert z \overset{\text{vis}}{\sim} x_m\}\text{: } \mathcal{O}(y;P_{m-1}) < k-1$$
where $k$ is the target order of visibility. Then
$$x_{m+1} \in \argmax_{x \in \Omega_{\text{free}}} \mathcal{G}_\text{naïve}(x; P_{m-1}).$$
\end{prop}

This shows that when the target order $k$ has been reached nowhere, then the next sensor will be chosen from the same pool as the previous one. 
Since this condition is always satisfied in the beginning if $k \geq 2$ and it is rare that two sensors have equal gain, the sensor will be chosen repeatedly. This will yield the trivial solution which places $k$ sensors in the same location. In most applications however, this solution is not desirable. It provides little robustness towards adversarial attacks and does not improve confidence in the depth sensors since it does not yield additional measurements from different angles.

We therefore use the following objective function which enforces distance between sensors.
\begin{equation}\label{eqn:gain_function}
    \mathcal{G}_k(x; P_n) := \min_{i=1,\cdots,n} \Vert x - x_i \Vert_p~ \mathcal{G}^\text{naïve}_k(x; P_n),
\end{equation}
where $||\cdot||_p$ is the $\ell_p$ norm of a vector in $\mathbb{R}^d$ and $P_n = \{x_1,...,x_n\}$. In this paper, $d=2$ and $p=2.$

\begin{algorithm}[H]
\caption{$\epsilon$-greedy algorithm}\label{alg:eps-greedy}
\begin{algorithmic}
\State $P = \text{list}()$
\While{$\int_{\Omega_{\text{free}}} \min(\mathcal{O}(y; P), k) dy < \delta k \lambda(\Omega_{\text{free}})$}
\State calculate $\mathcal{G}_k(x, P)$ for every $x \in \Omega_{\text{free}}$
\State calculate $M = \max_{z \in \Omega_{\text{free}}} \mathcal{G}_k(z, P)$
\State choose $x^*$ randomly from the set
\State $\{y \in \Omega_{\text{free}} \vert \mathcal{G}_k(y, P) \geq (1-\epsilon)M\}$
\State P.append($x^*$)
\EndWhile
\end{algorithmic}
\end{algorithm}

We present Algorithm~\ref{alg:eps-greedy}, which describes an $\epsilon$-greedy method to generate a sequence of sensors stored in $P$. If $\epsilon = 0$ we just call it the greedy algorithm. $k$ describes the target order of visibility and $\delta$ describes the termination threshold percentage of $\Omega_{\text{free}}$ (or $\Omega_{\text{vis}}$).
 However this is still costly to evaluate $\mathcal{G}_k(x,P)$ for every $x \in \Omega_{\text{free}}$.\\

For $k=1$, It is possible to derive a submodularity theory \cite{krause} to derive bounds that compare the relative performance of the greedy algorithm compared to the optimal solution; see \cite{ly2018greedy}.
 However when using the modified objective function $\mathcal{G}_k$ as an analog to the discrete derivative the submodular property is lost.

\subsection{Remark on computational complexity.}
Using the sweeping algorithm for graph environments from \cite{visalg} has a computational complexity of $\mathcal{O}(M^2)$ for the evaluation of $\mathcal{G}_k$. Thus using the $\epsilon$-greedy algorithm, placing $K$ sensors on an $M \times M$ grid in $\Omega_{\text{free}} \subset \mathbb{R}^2$ has complexity of $\mathcal{O}(KM^2)$. We can observe that most of the complexity stems from the evaluation of the gain function. While this is much faster than a brute-force search of sensor sets, the evaluation of the gain function can still be costly for large environments.

\section{GREEDY ALGORITHMS AIDED BY DEEP LEARNING}
Above we explained that the computational cost associated with each greedy step for adding a new sensor remains notably high. To address this challenge, we present a  deep learning approach in this section, building upon the strategy introduced in \cite{explorationsurveilance}, with the aim of significantly improving computational speed. 

Our method involves a neural network, denoted as $\mathsf{G}_{k,\theta}[\mathsf{X_0};\mathsf{D}_N ]$, trained using the data set $\mathsf{D}_N$. This neural network, trained with a suitable data set $\mathsf{D}_N$, approximates 
the gains $\mathcal{G}_k(x_{i,j})$ for $x_{i,j}$ on a grid over the environment, 
 thereby enhancing computational efficiency. 
 We will delve into the critical aspects of "generating" a performing training data set $\mathsf{D}_N,$ essential to the success of our approach.

\subsection{Network architecture}
We used the network architecture called TiraFL  described in \cite{jégou2017layers}.
For the order $k=1$ visibility case, it was shown in \cite{explorationsurveilance} that a UNet architecture is capable of approximating the gain function. However in our experiments a UNet architecture of similar size performed poorly for higher orders of visibility $k > 1$. 

\subsection{Training data}
As in the supervised learning setup, the data points in our study take the form of data pairs:
\( (\mathsf{X}, \mathsf{Y}).\). The input data $\mathsf{X}$ is composed of two parts, namely $\mathsf{X}\text{obs}$ and $\mathsf{X}\text{cumvis}$. \\
 
\textbf{$\mathsf{X}\text{obs}$} represents the "obstacle/environment map". For the data generation we used Massachusetts building footprint maps from \cite{MnihThesis} (The labels specifically). From this data set, we used random crops and a flood fill algorithm to assign random heights (between 0 and 1) to each distinguishable building to extend it to a 3D obstacle map. For these maps, we do not allow placements of sensors on top of buildings.\\
 
\textbf{$\mathsf{X}\text{cumvis}$} is the "cumulative visibility map," which provides the coverage information of a given set of sensors, and requires simulating the greedy algorithms offline, starting from a random initialization of the first sensor. Since 
$$y \overset{\text{vis}}{\sim} x = (x_1,...,x_n) \iff g_y(x_1,...,x_{n-1}) < x_n,$$
we see that $$\mathcal{O}_\text{vis}(y;x^1,x^2,...,x^n) = \ell \iff$$ 
$$\ell\text{-min} \{g_{x^i}(y_1,...,y_{n-1}) \text{ : } i=1,...,N\} < y_n,$$
where $\ell\text{-min}$ returns the $\ell$-th smallest element in the given set. 
For convenience, we define $C_\ell(y) = \ell\text{-min} \{g_{x^i}(y_1,...,y_{n-1}) \text{ : } i=1,...,N\}.$
Using this representation we can derive a simple update rule for $C_\ell$ when adding a new sensor $x \in \Omega_{\text{free}}$: 
$$C_\ell(y) \leftarrow \ell\text{-min} \{C_1(y), ..., C_k(y), g_x(y_1,...,y_{n-1})\}.$$

The output data \textbf{$\mathsf{Y}$} is the image of $\mathcal{G}_k$ on a grid overlaying the environment described by $\mathsf{X}\text{obs}$. Further both the input and gain are normalized such that each element is in $[0,1]$. Depending on the specific greedy algorithm used to generate $\mathsf{X}\text{cumvis}$ (and consequently $\mathsf{Y}$), we denote the corresponding datasets as $\mathsf{D}_{N_1}^0$ for the pure greedy approach and $\mathsf{D}_{N_2}^\epsilon$ for the $\epsilon$-greedy method. \\


\begin{table}[H]
 \caption{Data sets used in our study.}
    \label{tab:data_sets}
    \centering
 \begin{tabular}{|l | l | l| l |}
    \hline
      & $\mathsf{D}^0_N$ & $\mathsf{D}^\epsilon_N$ & $\mathsf{D}^+_N$ \begin{minipage}[c][6mm][t]{0.1mm}\end{minipage} \\
      \hline
      \begin{minipage}[b][3mm][t]{0.1mm}\end{minipage}N & 6531 & 2511 & 6342 (3831+2511) \\
      $\epsilon$ & $0$ & $0.05$ & $0$ and $0.05$\\
      Grid & $128 \times 128$ & $128 \times 128$ & $128 \times 128$\\
      Order & 2 & 2 & 2 \\
      \hline
\end{tabular}
\end{table}
\noindent $\mathsf{D}^0_N$ was generated using the greedy algorithm with $\epsilon = 0$. $\mathsf{D}^\epsilon_N$ was generated using the $\epsilon$ greedy algorithm with $\epsilon = 0.05$. Notably, $\mathsf{D}_{N}^+ = \mathsf{D}_{N_1}^0 \cup \mathsf{D}_{N_2}^\epsilon$ combines both types of data for comprehensive analysis and comparison. The merged data set only contains a subset of $\mathsf{D}^0_N$ to keep the number of data points in $\mathsf{D}_{N}^+$ and $\mathsf{D}_{N_1}^0$ at a similar level.  We will see, in Section~\ref{sec:geometry}, that this combined dataset leads to performing neural networks. 
 
 \section{EFFECT OF DATA GEOMETRY}\label{sec:geometry}
In this section, we analyze the networks trained with $\mathsf{D}^0_N$ and $\mathsf{D}^+_N$ in an urban environment. Accurate results can be observed we looking at test data, however, this only checks the accuracy of one step. As the network is the approximation of the gain function, it needs to be applied multiple times where its input depends on the output of the previous iteration. Initially, the network's predictions are accurate, but its reliability diminishes as the simulation progresses. This decrease in accuracy typically occurs when the network encounters a scenario where two or more sensors, which are far apart, produce a similar gain. Due to approximation error, the network may choose a sub-optimal option in these cases, resulting in decreased accuracy for future placements.

Note that the gain function from Equation \eqref{eqn:gain_function} cannot be evaluated when no sensors have been placed. Hence, we fix the first sensor at $(0,0)$. Then we generate the subsequent sensors using the methods discussed above. As a reference, we will use the greedy algorithm where the function $\mathcal{G}_k$ is calculated and the maximal value is chosen.

\begin{figure}[H]
    \begin{subfigure}[b]{0.475\linewidth}
        \centering
        \includegraphics[width=\linewidth]{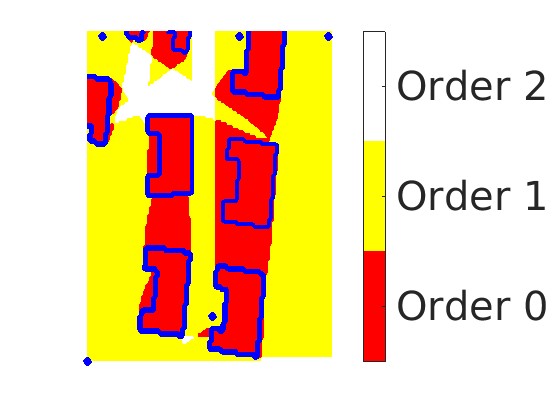}
        \caption{5 sensors, $\mathsf{D}^0_N$}    
    \end{subfigure}
    \hfill
    \begin{subfigure}[b]{0.475\linewidth}  
        \centering 
        \includegraphics[width=\linewidth]{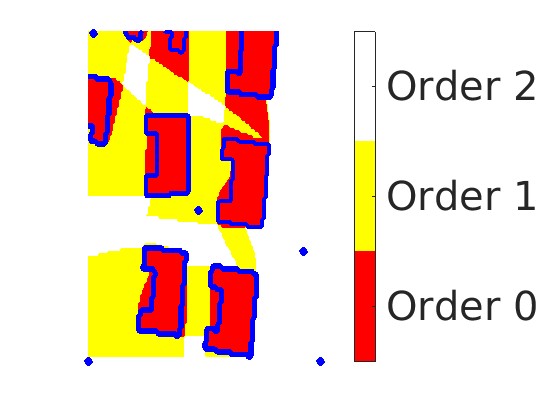}
        \caption{5 sensors, $\mathsf{D}^+_N$}
    \end{subfigure}
    \vskip\baselineskip
    \begin{subfigure}[b]{0.475\linewidth}   
        \centering 
        \includegraphics[width=\linewidth]{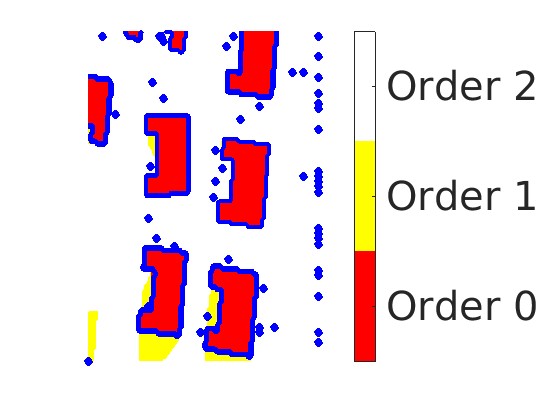}
        \caption{60 sensors, $\mathsf{D}^0_N$}
    \end{subfigure}
    \hfill
    \begin{subfigure}[b]{0.475\linewidth}   
        \centering 
        \includegraphics[width=\linewidth]{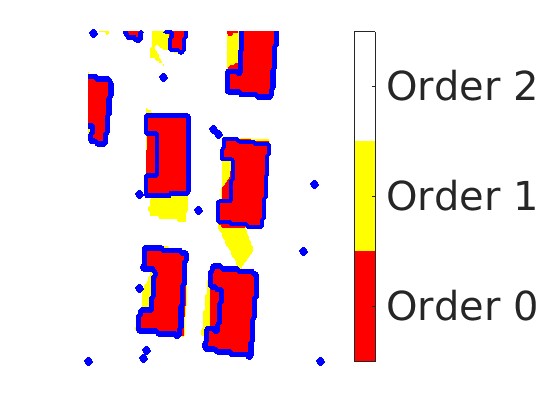}
        \caption{13 sensors, $\mathsf{D}^+_N$}
    \end{subfigure}
    \caption{Comparison of order of visibility function when using the neural network trained with $\mathsf{D}^0_N$ and $\mathsf{D}^+_N$. The first row shows the result after placing 5 sensors. The second row show the result after the threshold volume was reached.}
    \label{fig:net_results}
\end{figure}

In Figure~\ref{fig:net_results}, we see that both neural networks show promising results for the first 5 sensors. However, as shown in the bottom row for later iteration the neural network trained with $\mathsf{D}^0_N$ predicts close points repeatedly. This can be observed by the accumulation of blue dots (sensors) on the right side. The network trained with $\mathsf{D}^+_N$ however performs well throughout the whole algorithm and achieves target volume after placing 13 sensors. 

\begin{figure}[H]
    \centering
    \includegraphics[width = \linewidth]{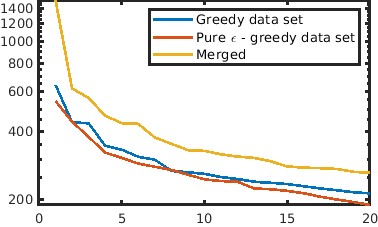}
    \caption{Comparison of the first 20 singular values of data sets in final stages of placement. The x-axis depicts the number of the singular values ordered by their magnitude and the y-axis depicts the value of the singular values.}\label{fig:singular-value-comp}
\end{figure}

To gain further insight, we delve into the data geometry of both data sets, with a particular focus on later stages in the simulation. We hope to shed light on the mechanisms that underlie the improved performance of the extended data set.
To conduct this analysis, we initially extracted data points from the training data set that corresponded to either the final 10 sensors placed or the second half of the data when fewer than 20 points were required to reach the visibility threshold. This data set is then centered around its mean and PCA is applied to extract the principle directions and the corresponding singular values. We then compare the singular values to get insight into the data geometries in Figure~\ref{fig:singular-value-comp}.

\vspace{-1.5em}
The singular values describe the variances of the data set in the associated principal (singular) direction. While the dimensionality of the Greedy and $\epsilon$-greedy data sets are similar, the merged data set is higher dimensional. This may explain the bad predictions in the later stages of the simulation for the greedy data set.

\section{DEMONSTRATION}
\begin{figure}[H]
     \centering
     \includegraphics[width = 0.8\linewidth]{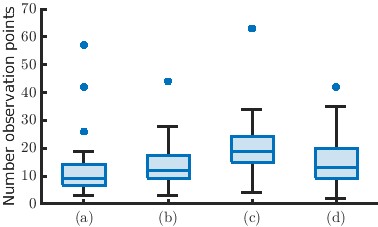}
     \caption{Distribution of the number of sensors needed for 54 urban obstacle maps. Shows the results for (a) $\mathcal{G}_k$ with $\epsilon = 0$. (b) neural network trained with $\mathsf{D}_{N}^+$. (c) Random sensor placement. (d) neural network trained with $\mathsf{D}_{N_1}^0$}\label{fig:num-sensors-comp}
 \end{figure}

With the trained network, the computation for each greedy step is 105 times faster when using 64 cores on an AMD EPYC 7763 CPU for the greedy algorithm and a NVIDIA A100 GPU for the network approximation. In Figure~\ref{fig:num-sensors-comp}, we compare the distributions of the number of sensors needed to reach a prescribed threshold.

The neural network trained with $\mathsf{D}_{N}^+$ produces results comparable to using calculation of the function $\mathcal{G}_k$. Thus we can use the neural network to efficiently approximate the gain function which allows for a fast application of the greedy algorithm for 3D urban maps.

\begin{figure}[H]
    \centering
    \includegraphics[width = \linewidth]{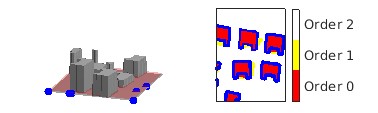}
    \caption{Horizontal slice}
    \label{fig:hor-slice}
    \includegraphics[width = \linewidth]{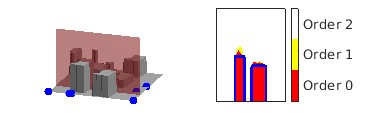}
    \caption{Vertical slice}
    \label{fig:ver-slice}
\end{figure}
We present the final results of the greedy algorithm using the neural network in Figures~\ref{fig:hor-slice}-\ref{fig:ver-slice}. The red plane on the left represents slices of the order of the visibility map shown on the right.

\section{Summary}
We propose a greedy algorithm for placing a minimal set of sensors that satisfies a multi-coverage requirement. We show that the framework can be well approximated using deep learning techniques for a broad class of obstacles, which significantly improves computation time. The geometric properties of the data used for training play an important role in the accuracy and variability in generating the training data is crucial for the success of the algorithm. We envision that the insight into data geometry has large implications for neural network approximation tasks.

\section{Acknowledgement}
Taus is supported by Army Research Office, under Cooperative Agreement Number W911NF-19-2-0333. Tsai is partially supported by Army Research Office, under Cooperative Agreement Number W911NF-19-2-0333 and National Science Foundation Grant DMS-2110895.




\end{document}